\documentclass[letterpaper]{article} 
\usepackage[preprint]{aaai2027}  
\usepackage[hyphens]{url}  
\usepackage{graphicx} 
\usepackage{natbib}  
\usepackage{caption} 
\usepackage{algorithm}
\usepackage{algorithmic}

\usepackage{newfloat}
\usepackage{listings}
\DeclareCaptionStyle{ruled}{labelfont=normalfont,labelsep=colon,strut=off} 
\floatstyle{ruled}
\newfloat{listing}{tb}{lst}{}
\floatname{listing}{Listing}

\usepackage{booktabs}

\title{MTLiquid: Enabling Efficient Multi-Task Learning using Liquid Neural Networks for Lightweight Healthcare Monitoring Systems}
\author{
    Rachmad Vidya Wicaksana Putra,
    Fahad Abdul Rauf,
    Muhammad Shafique
}
\affiliations{
    eBRAIN Lab, New York University (NYU) Abu Dhabi, Abu Dhabi, United Arab Emirates \\
    \{rachmad.putra, fa2788, muhammad.shafique\}@nyu.edu 

}

\usepackage{xcolor}
\usepackage{amsmath}
\usepackage{amssymb}
\usepackage{booktabs}
\usepackage{multirow}
\usepackage{tikz}

\begin{document}

\maketitle

\begin{abstract}
Continuous-time sensing and monitoring with timely and accurate decision-making are critical for many real-world applications. 
In healthcare monitoring systems, physiological signals are often available or sampled at irregular time intervals, hence requiring continuous-time processing to provide accurate prediction. 
Moreover, such systems often need to solve multiple detection/prediction tasks to provide a comprehensive patient review from different physiological aspects for more accurate decision-making. 
To solve this, continuous-time neural networks (CTNNs) can be employed. 
However, state-of-the-art works typically solve only one task at each network, thereby limiting their efficiency gains.  
To address this limitation, we propose \textbf{\textit{MTLiquid}}, \textit{a novel methodology to enable efficient multi-task learning in continuous-time processing for healthcare monitoring systems through effective network design and training strategy.}
MTLiquid employs: (1) multiple input and output heads to accommodate different tasks, while sharing the same backbone network across tasks; as well as (2) an effective training strategy that leverages a loss-weighting technique to balance learning updates across different tasks and a proportional data presentation technique to address imbalanced dataset sizes.  
Experimental results for mortality prediction (P12) and sepsis early detection (P19) tasks for ICU patients show that, MTLiquid achieves strong performance (AUROC: 0.84 for P12 and 0.94 for P19) comparable to the state-of-the-art single-task learning in both continuous-time networks (AUROC: 0.84 for P12 and 0.95 for P19) and discrete-time networks (AUROC: 0.79-0.82 for P12 and 0.92-0.94 for P19), while incurring significantly smaller memory cost by 44\%-94\%.  
These results highlight the potential of our MTLiquid methodology to enable lightweight continuous-time healthcare monitoring systems for better decision-making.
\end{abstract}


\section{Introduction}
\label{Sec_Intro}

Continuous-time sensing and monitoring with timely and accurate decision-making are very important in many real-world application use-cases, especially in healthcare systems.  
For instance, healthcare monitoring systems often collect physiological data at irregular time intervals, such as data from patients in the Intensive Care Unit (ICU) related to in-hospital mortality~\cite{silva2012predicting} and sepsis~\cite{reyna2020early}.
Hence, accurate mortality prediction and sepsis early detection are the key to making timely and accurate decisions in devising proper treatments for ICU patients. 
Furthermore, solving multiple tasks (i.e., mortality prediction and sepsis early detection) provides a comprehensive patient review for more accurate decision-making.
Employing a conventional discrete-time processing approach, such as \textit{Deep Neural Networks (DNNs)}~\cite{Ref_LeCun_DeepLearning_Nature15}, \textit{Spiking Neural Networks (SNNs)}~\cite{Ref_Putra_FSpiNN_TCAD20, Ref_Putra_SpikeNAS_TAI26}, as well as \textit{Recurrent Neural Network (RNN)} and \textit{Long Short-Term Memory (LSTM)} models~\cite{hochreiter1997lstm, cho2014gru, Ref_Ghojogh_RNNnLSTM_arXiv23, Ref_Liu_SurveyRNNs_arXiv25}, to handle continuous-time data can only provide sub-optimal performance due to its limitations in capturing the significance of continuous-time information in neural behavior. 
This highlights that a continuous-time processing approach is required.
Additionally, mortality prediction and sepsis early detection tasks in healthcare monitoring systems require continuous monitoring using wearable devices with strict/tight memory and power budgets. 

\textbf{Targeted Research Problem}: \textit{How can we perform simultaneous multi-task learning using the continuous-time processing approach for healthcare monitoring systems (i.e., mortality prediction and sepsis early detection)?} 
An efficient solution to this problem may enable efficient healthcare systems capable of solving multiple tasks with a single shared network model.

\subsection{State-of-the-Art and Their Limitations}

To address the targeted problem, several works have been proposed in the literature to process irregularly-sampled data, support continuous-time series processing, and/or accommodate multi-task learning capabilities, as described below.
\begin{itemize}
    \item Prior approaches to irregularly-sampled data often employ discrete-time recurrent network backbones (e.g., RNN or LSTM architectures) used in the single-task setting. 
    To achieve this, they handle irregularity by adding elapsed time or a decay-weighted hidden state as an auxiliary input~\cite{che2018recurrent, cao2018brits, shukla2021mtand}. 
    Such networks may be extended to handle multiple tasks by attaching separate output heads to a shared recurrent backbone~\cite{caruana1997multitask, futoma2017sepsis, harutyunyan2019multitask}. 
    However, the recurrent cell still updates on a fixed discrete schedule regardless of the true inter-arrival time of data sampling. 
    \textit{Consequently, they often offer sub-optimal performance.}
    \item To effectively handle irregularly-sampled data and perform continuous-time series processing, state-of-the-art works consider \textit{Liquid Neural Networks (LLNs)} as the prominent approach, which include \textit{Liquid-Time Constant (LTC)}~\cite{hasani2021liquid} and \textit{Closed-form Continuous-time (CfC)}~\cite{hasani2022closed} networks.
    The LTC model is based on differential equation-based neurons interconnected via sigmoidal synapses.
    However, its real-world applicability is limited by its requirement for a numerical differential equation solver, which slows down the processing time for both training and inference. 
    This limitation is alleviated by the CfC model through approximation using the closed-form solution of the differential equation~\cite{hasani2021liquid}.
    However, these state-of-the-art works primarily focus on developing high performance/quality solutions for continuous-time series problems.
    \textit{Hence, studies in employing LNNs for solving multi-task problems have not been explored.}
\end{itemize}
The above discussion highlights that, \textit{in general, state-of-the-art works have not explored the 
potential of continuous-time series processing for multi-task learning.}
Consequently, their benefits for solving multi-task prediction/detection (i.e., mortality prediction and sepsis early detection) for healthcare monitoring systems have not been investigated.

\smallskip
\textbf{Associated Research Challenges:}
Enabling continuous-time series processing under a multi-task learning scenario for healthcare monitoring systems is non-trivial, as it imposes the following critical research challenges.
\begin{itemize}
    \item The solution should employ an efficient network architecture suitable for continuous-time healthcare monitoring, especially for solving the targeted tasks (i.e., mortality prediction and sepsis early detection).
    \item The solution should employ an effective multi-task learning mechanism so that the network model can address any given task within the targeted ones without significant performance/accuracy degradation compared to the single-task scenario. 
    \item The solution should consider a compensation mechanism to address potentially imbalanced dataset sizes from different tasks that will be incorporated into the multi-task training process. 
\end{itemize}

\subsection{Our Novel Contributions}

To address the targeted problem and its related challenges, we propose \textbf{\textit{MTLiquid}}, \textit{a novel methodology to enable efficient multi-task learning using continuous-time series processing for mortality prediction and sepsis early detection in healthcare monitoring systems.}
This is also the first work that studies the potential of LNNs for solving multi-task learning problems. 
The MTLiquid methodology employs the following novel contributions:
\begin{itemize}
    \item \textbf{Efficient Network Architecture Design:}
    It aims to design an efficient CfC-based network architecture for solving mortality prediction and sepsis early detection tasks.
    The designed network has multiple input and output heads to accommodate different tasks, while sharing the same backbone network across tasks.
    \item \textbf{Effective Training Strategy:}
    It aims to devise a training strategy that leverages: (1) a loss-weighting technique to balance learning updates across different tasks, and (2) a proportional data presentation technique to address imbalanced dataset sizes.
\end{itemize}
\textbf{Key Results:} 
To evaluate our MTLiquid methodology, we implement it using PyTorch and run it on a single Nvidia RTX 4090 Ti GPU device.
Experimental results for mortality prediction (P12) and sepsis early detection (P19) show that, MTLiquid achieves strong performance (AUROC: 0.84 for P12 and 0.94 for P19) comparable to the state-of-the-art single-task learning in both continuous-time networks (AUROC: 0.84 for P12 and 0.95 for P19) and discrete-time networks (AUROC: 0.79-0.82 for P12 and 0.92-0.94 for P19), while saving the memory cost by 44\%-94\%.  
These results highlight the potential of MTLiquid methodology to enable lightweight continuous-time healthcare monitoring systems. 

\section{Background}
\label{Sec_Back}

\subsection{Liquid Neural Networks (LNNs)}
\label{Sec_Back_LNNs}

LNNs allow for continuous-time recurrence by defining each neuron's hidden state as the solution of a nonlinear ordinary differential equation (ODE) ~\cite{chen2018neuralode}, in contrast to the fixed-interval discrete update rules used by RNNs and LSTMs. 
An LTC network~\cite{hasani2021liquid, lechner2020ncp} is the foundational model in this class, modeling each neuron as a leaky integrator whose time constant is input-dependent rather than fixed. 
Since the gating term in LTC models depends on both the current state and the input, the effective time constant for each neuron varies continuously with the input signal at inference time, rather than being fixed at initialization.
Two properties of this formulation are directly relevant to irregularly-sampled data time series, as described below. 
\begin{itemize}
    \item Since the neuron state is defined over continuous time, the network can compute a valid state at arbitrary elapsed intervals between observations, without requiring a decay-weighted approximation of missing sampled data at irregular time, similar to GRU-D-style approaches \cite{che2018recurrent, de2019gruodebayes}. 
    \item The bounded gating function together with a strictly positive baseline time constant ensures that the coefficient which governs state decay remains strictly negative for all inputs. This ensures bounded and stable state trajectories independent of input magnitude or elapsed interval.
\end{itemize}

The main limitation of this formulation is associated with its relatively slow computation, since evaluating the neuron state requires a numerical ODE solver at every forward pass, at both training and inference time. 
This introduces a sequential multi-step loop per timestep, substantially increasing training and inference latency. 
This limitation motivates a closed-form approximation-based approach that leads to the development of the CfC solution, which will be discussed in the next sub-section.

\subsection{Closed-form Continuous-time Neural Networks (CfCs)}
\label{Sec_Back_CfCs}

CfCs address the limitation of ODE-solver for LTCs by deriving a closed-form approximation to the LTC state update, so that the recurrent cell does not need a solver to compute its update at each timestep, during both training and inference phases. 
The derivation begins from the LTC state, as expressed in Equation~\ref{Eq_LTC}.

\begin{equation}
  \begin{split}
  \frac{dx(t)}{dt} = & -\left[\frac{1}{\tau} + f(x(t), I(t); \theta)\right] x(t) + \\ 
  & f(x(t), I(t); \theta) A
  \label{Eq_LTC}
  \end{split}
\end{equation}
\smallskip

\noindent where $x(t)$ is the neuron state, $I(t)$ is the input, $\tau$ is the time-constant parameter, $A$ is the bias, and $f(\cdot;\theta)$ is the neural network that gates both the effective decay rate and the steady-state target of the neuron. 
Under a piecewise-constant input assumption, Equation~\ref{Eq_LTC} has a closed-form approximation with a tight error bound, which means that the ODE does not need to be solved numerically. 
This yields the CfC hidden-state update, which can be expressed as Equation~\ref{Eq_CfC}.

\begin{equation}
    \begin{split}
    h(t) = {}& \sigma(-f(x, I; \theta_f)\, t) \odot g(x, I; \theta_g) \\
    & + \left[1 - \sigma(-f(x, I; \theta_f)\, t)\right] \odot h(x, I; \theta_h)
    \label{Eq_CfC}
    \end{split}
\end{equation}
\smallskip

\noindent where $f(\cdot;\theta_f)$, $g(\cdot;\theta_g)$, and $h(\cdot;\theta_h)$ are single-hidden-layer network heads, $\sigma$ is the sigmoid function, $\odot$ is the Hadamard product, and $t$ is the elapsed time since the last observation, supplied as an explicit input rather than a fixed step size. 
The $\sigma(-f(\cdot)\,t)$ term acts as a time-continuous gate that inserts between the two heads $g(\cdot)$ and $h(\cdot;\theta_h)$. 
Since $f(\cdot;\theta_f)$ depends on the input, the gate's decay rate is not fixed, but changes based on what the network sees at each step (\textit{liquid)}.
Here, $t$ appears directly in Equation~\ref{Eq_CfC} rather than requiring integration. 
Therefore, computing $h(t)$ can be done a single forward pass, giving approximately an order-of-magnitude speedup over ODE-based LTCs. 
Irregular time gaps do not need special handling and $t$ is simply another input to the gate, which is why CfCs are well suited to irregularly sampled time series.

\section{The MTLiquid Methodology}
\label{Sec_MTL}

\subsection{Overview}
\label{Sec_MTL_Overview}

Our MTLiquid methodology aims to enable efficient multi-task learning with CfC-based continuous-time series processing for mortality prediction and sepsis early detection through efficient network architecture design and effective training strategy.
To achieve this, we first consider $K$ classification tasks defined over irregularly-sampled clinical data time series, where each task $k$ has its own observation space, sampling density, and missing-data pattern. 
For task $k$, an instance is a sequence of $T_k$ timestamps $\{\tau_1, \dots, \tau_{T_k}\}$, an observation matrix $X^{(k)} \in \mathbb{R}^{T_k \times D_k}$, and a binary missing-data mask $M^{(k)} \in \{0,1\}^{T_k \times D_k}$ indicating which of the $D_k$ channels are actually measured at each $\tau_i$. 
Rather than treating $M^{(k)}$ as an incidental artifact of the data, MTLiquid treats it as a first-class data input, i.e., whether a measured variable is informative in clinical data time series, since sampling frequency is often driven by acuity~\cite{che2018recurrent, lipton2016missing}.

\begin{figure*}[t]
\centering
\includegraphics[width=\textwidth]{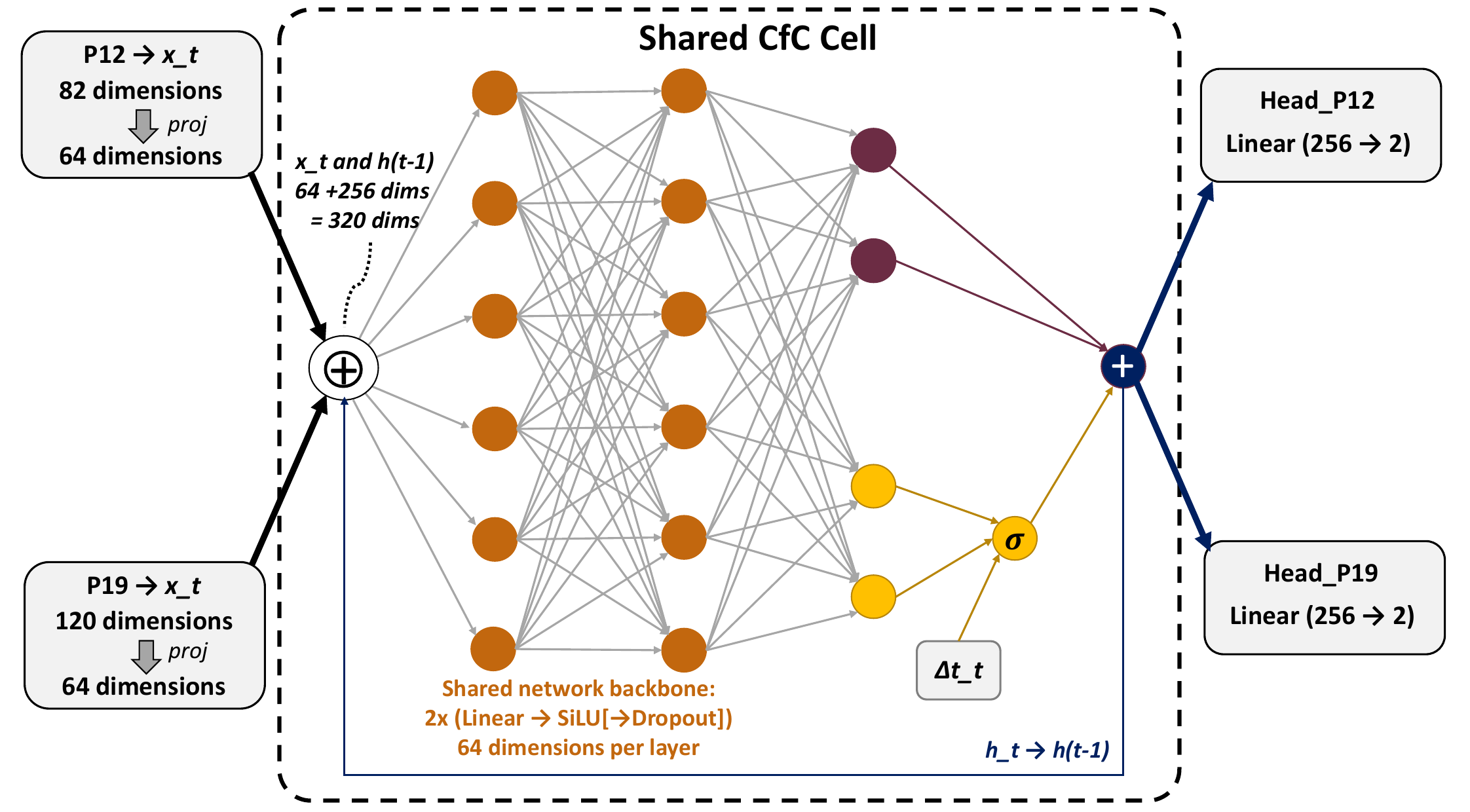}
\caption{Multi-task architecture in MTLiquid based on the CfC concept. 
Task-specific inputs are projected to a common 64-dimensional space, concatenated with $h(t-1)$, and passed through a shared backbone.
Candidate and time-gate branches (gated by $\sigma$ over $\Delta t_t$) update the hidden state, which feeds task-specific linear heads.}
\label{fig:cfc-arch}
\end{figure*}

\subsection{Efficient Network Architecture Design}
\label{Sec_MTL_NetArch}

In this work, the targeted tasks include the mortality prediction (i.e., so-called task-P12)~\cite{silva2012predicting} and the sepsis early detection (so-called task-P19)~\cite{reyna2020early}.
Therefore, we design the network architecture specifically to solve these tasks.
To achieve this, we design a single model with parameter-sharing for different tasks, where it maps each task's heterogeneous and irregularly-sampled input data to a task-specific prediction~\cite{caruana1997multitask, zhang2021mtlsurvey, harutyunyan2019multitask, Ref_Putra_MTSpark_arXiv24, Ref_Putra_SwitchMT_arXiv25}, while following some principles below.
\begin{enumerate}
    \item Sharing most of the network model parameters across tasks to optimize the model size.
    \item Letting elapsed time directly affect the evolution of the hidden state, thereby giving our architecture the ``liquid'' characteristics.
\end{enumerate}

We first define how task-specific inputs are brought into a shared network backbone since the input dimension differs across tasks. 
Each task is given its own projection $x_t^{(k)} = W_k \hat{x}_t^{(k)} + b_k$, $W_k \in \mathbb{R}^{d \times D_k^{\text{in}}}$, into a common embedding dimension $d$. 
In our example in Figure~\ref{fig:cfc-arch}, it has 64 dimensions. 
This is the only task-specific transformation applied before the recurrent network backbone, because from this point on, sequences for task-P12 and task-P19 are indistinguishable in shape, and they are processed by identical shared weights. 
This is an intentional design choice to minimize the number of weight parameters. 
Furthermore, it isolates task variance to the boundary of the network (i.e., only input projection and output head) and forces all cross-task sharing to happen inside a single continuous-time network backbone. 

Raw channels from each task are first forward-filled, similar to a GRU-D-style scheme~\cite{che2018recurrent}.
Specifically, at step $t$, a missing entry in channel $d$ is replaced with the most recently observed value in that channel, rather than a fixed placeholder such as zero. 
Concretely, letting $x_{t,d}$ denote the raw (possibly missing) reading/sampling and $m_{t,d} \in \{0,1\}$ its mask bit, then the forward-fill policy can be expressed as Equation~\ref{eq:ffill}.

\begin{equation}
\hat{x}_{t,d} = m_{t,d}\, x_{t,d} + (1-m_{t,d})\,\hat{x}_{t-1,d},
\label{eq:ffill}
\end{equation}
\smallskip

\noindent This policy carries forward per channel across the sequence, thereby providing a good estimate of what the channel's true value is in the area of missing data. 
The imputed value is concatenated with its own mask bit before passed to the model. 
This yields a per-timestep input dimensionality, i.e., 82 dimensions for task-P12 (41 imputed values concatenated with 41 mask bits) and 120 dimensions for task-P19 (60 imputed values concatenated with 60 mask bits), as shown in Figure~\ref{fig:cfc-arch}.

The projected sequence $x_t^{(k)}$, together with the elapsed time $\Delta t_t^{(k)} = \tau_t^{(k)} - \tau_{t-1}^{(k)}$, is consumed by a single shared CfC cell and applied identically regardless of task; see the pseudo-code in Algorithm~\ref{alg:cfc-cell}. 
Following the closed-form solution in Equation~\ref{Eq_CfC}, the cell computes two candidate hidden states from the current input and previous state, as expressed in Equation~\ref{eq:candidates}.

\begin{equation}
g_t = \mathrm{ff}_1(x_t^{(k)}, h_{t-1}), \qquad h_t^{\text{cand}} = \mathrm{ff}_2(x_t^{(k)}, h_{t-1})
\label{eq:candidates}
\end{equation}
\smallskip

\noindent This Equation~\ref{eq:candidates} corresponds to $g(x,I;\theta_g)$ and $h(x,I;\theta_h)$ in Equation~\ref{Eq_CfC}. 

The gate itself instantiates $f(x,I;\theta_f)$ as a linear function of the backbone features rather than a single unconstrained network head: $f(x_t^{(k)}, h_{t-1}; \theta_f) = t_a(x_t^{(k)}, h_{t-1})$, with a learned bias $t_b(x_t^{(k)}, h_{t-1})$ additionally shifting the gate's zero-crossing independently of $\Delta t$,

\begin{equation}
t_{\text{interp}} = \sigma\!\left(-\underbrace{t_a(x_t^{(k)}, h_{t-1})}_{f(x,I;\theta_f)} \cdot \Delta t_t^{(k)} + t_b(x_t^{(k)}, h_{t-1})\right)
\label{eq:gate}
\end{equation}
\smallskip

\noindent reducing to the gate of Equation~\ref{Eq_CfC} exactly when $t_b \equiv 0$. The two candidates are then combined as in Equation~\ref{Eq_CfC}, and can be expressed as Equation~\ref{eq:hidden_update}. 

\begin{equation}
h_t = g_t \odot t_{\text{interp}} + h_t^{\text{cand}} \odot (1 - t_{\text{interp}}).
\label{eq:hidden_update}
\end{equation}
\smallskip

Because $t_a$ and $t_b$ are functions of the input rather than fixed constants, the effective time-constant of the recurrence adapts per-step and per-channel-pattern, while remaining a single closed-form evaluation. Therefore, no ODE solver is invoked at train or inference time. 
This is the same $\mathrm{ff}_1, \mathrm{ff}_2, t_a, t_b$ across both tasks.
This means the LTC dynamics that decide how aggressively to trust new evidence given the observed $\Delta t$, are actually a shared network. 
This directly tests our hypothesis, i.e., a single continuous-time state-transition function can simultaneously serve tasks with different sampling irregularity (i.e., task-P12 and task-P19 have different sampling profiles).

Output data from the final hidden state $h_T^{(k)}$ are passed to a task-specific linear head, $\hat{y}^{(k)} = W_{\text{head},k} h_T^{(k)} + b_{\text{head},k}$, producing per-task classification. 
This keeps the number of parameters of the shared network backbone essentially independent of the number of tasks. 
Consequently, adding a task only costs one input projection and one linear output head, with the same shared recurrent network backbone.

\begin{algorithm}[t]
\caption{Shared CfC Cell ($f_{\text{shared}}$)}
\label{alg:cfc-cell}
\begin{algorithmic}[1]
\REQUIRE $z$ : projected input at current step, dim $d$; \\
$h_{\text{prev}}$ : recurrent state from previous step, dim $H$; \\
$\Delta t$ : elapsed time since last observation; \\
$W_1, b_1$ : backbone layer 1 weights/bias; \\
$W_2, b_2$ : backbone layer 2 weights/bias; \\
$W_{\text{ff1}}, W_{\text{ff2}}$ : candidate branch weights; \\
$W_{t_a}, W_{t_b}$ : time-gate branch weights;
\STATE $u \leftarrow \text{concat}(z, h_{\text{prev}})$; //merge input and recurrent state, dim $d+H$
\STATE $u_1 \leftarrow \text{SiLU}(W_1 u + b_1)$; // backbone layer 1
\STATE $b \leftarrow \text{SiLU}(W_2 u_1 + b_2)$; // backbone layer 2, shared across P12/P19
\STATE $\text{ff}_1 \leftarrow \tanh(W_{\text{ff1}} \, b)$; // candidate branch 1
\STATE $\text{ff}_2 \leftarrow \tanh(W_{\text{ff2}} \, b)$; // candidate branch 2
\STATE $t_a \leftarrow W_{t_a} \, b$; // time-gate branch, term 1
\STATE $t_b \leftarrow W_{t_b} \, b$; // time-gate branch, term 2
\STATE $\sigma_t \leftarrow \text{sigmoid}(t_a \cdot \Delta t + t_b)$; // gate value: how much of the elapsed interval to apply
\STATE $h \leftarrow (1 - \sigma_t) \cdot \text{ff}_1 + \sigma_t \cdot \text{ff}_2$; // closed-form interpolation between candidates
\STATE $h_{\text{prev}} \leftarrow h$; // carry state forward to next step
\STATE \textbf{return} $h$
\end{algorithmic}
\end{algorithm}

\subsection{Effective Training Strategy}
\label{Sec_MTL_Training}

\begin{algorithm}[t]
\footnotesize
\caption{Our Multi-Task Training Strategy}
\label{alg:mtliquid}
\begin{algorithmic}[1]
\REQUIRE $D_{p12}, D_{p19}$ : the two datasets P12 and P19; \\
$f_{shared}$ : shared recurrent cell; \\
$proj_{p12}, proj_{p19}$ : per-task input layers; \\
$head_{p12}, head_{p19}$ : per-task output layers; \\
$\sigma_{p12}, \sigma_{p19} = 0, 0$ : loss weights; \\
$N_{epochs}$, $R$ : max repeat for shorter loader; \\
\FOR{$epoch = 1$ to $N_{epochs}$}
    \FOR{$batch = 1$ to $B$}
        \STATE $batch_{12} \leftarrow (x_{12}, dt_{12}, mask_{12}, y_{12})$ from $D_{p12}$
        \STATE $batch_{19} \leftarrow (x_{19}, dt_{19}, mask_{19}, y_{19})$ from $D_{p19}$
        \FOR{$task \in \{p12, p19\}$}
            \STATE $(x, dt, mask, y) \leftarrow batch_{task}$
            \STATE $proj \leftarrow proj_{task}$, \; $head \leftarrow head_{task}$
            \STATE $\hat{x} \leftarrow 0$
            \FOR{$t = 1$ to $T$}
                \STATE $\hat{x} \leftarrow mask[t] \cdot x[t] + (1-mask[t]) \cdot \hat{x}$
                \STATE $z \leftarrow proj(\hat{x}, mask[t])$
                \STATE $h \leftarrow f_{shared}(z, h_{prev}, dt[t])$
                \STATE $h_{prev} \leftarrow h$
            \ENDFOR
            \STATE $h_{final} \leftarrow h$ from last observed $t$
            \STATE $\hat{y} \leftarrow head(h_{final})$
            \STATE $loss[task] \leftarrow \text{CE}(\hat{y}, y, w_{task})$
        \ENDFOR
        \STATE $total\_loss \leftarrow \dfrac{loss[p12]}{\sigma_{p12}^2} + \dfrac{loss[p19]}{\sigma_{p19}^2} + \log\sigma_{p12} + \log\sigma_{p19}$
        \STATE update all parameters using $total\_loss$
    \ENDFOR
    \STATE decay($lr$);  
    \STATE eval;
\ENDFOR
\end{algorithmic}
\end{algorithm}

To train the developed network architecture for task-P12 \cite{silva2012predicting} and task-P19 \cite{reyna2020early}, we propose an effective multi-task training strategy.
Its pseudo-code is presented in Algorithm~\ref{alg:mtliquid} and described in the following.

\smallskip
\textbf{Imbalanced Datasets:}
Task-P12 and task-P19 differ substantially in dataset size, so pairing one batch from each dataset per optimization step can lead to one of the following problems. 
\begin{itemize}
    \item it can starve the larger/longer dataset, if the data presentation to the network is capped to follow the smaller/shorter dataset's length, or 
    \item it forces the smaller/shorter dataset loader to repeat many times within a single epoch, if the data presentation to the network is stretched to match the larger/longer dataset's length, risking memorization and thus overfitting of the smaller task.
\end{itemize}
We address this with \textit{a capped-cycling schedule policy}: the smaller/shorter dataset loader is cycled, but repetition is bounded by a maximum-repeat factor, which we can personally tune depending on the nature of the datasets, 
and the larger/longer dataset loader is length-matched per epoch while still reshuffling on every pass. 
This keeps both tasks' gradient signal present in every optimization step and prevents any task loader from dominating or reducing into memorization.

\smallskip
\textbf{Loss Function:}
Rather than fixing the relative weight between $\mathcal{L}_{\text{P12}}$ and $\mathcal{L}_{\text{P19}}$ as a hyperparameter, we adopt the task-uncertainty formulation of Kendall, Gal, and Cipolla \cite{kendall2018multitask}, introducing one learned log-variance parameter-per-task $\log\sigma_k$, and optimizing

\begin{equation}
\mathcal{L} = \sum_{k=1}^{K} \frac{\mathcal{L}_k}{2\sigma_k^2} + \log \sigma_k 
\label{eq:kendall_loss}
\end{equation}
\smallskip

\noindent Each $\mathcal{L}_k$ is a class-weighted cross-entropy, with weights set to the inverse positive-class prevalence of that task's training split, so that the task's inherent label imbalance relative to the other tasks is handled automatically. 
The $\log\sigma_k$ regularization term prevents a task from being trivially down-weighted to zero contribution by driving $\sigma_k \to \infty$. 
As $\sigma_k$ is learned jointly with the network weights, the relative influence of task-P12 and task-P19 on the shared backbone is allowed to shift over training as the two tasks' relative difficulty changes, rather than being fixed at initialization.

Alternative strategies for balancing task losses include gradient normalization, which rescales per-task gradient magnitudes to a common target norm \cite{chen2018gradnorm}; gradient surgery, which projects away conflicting components between task gradients before the update \cite{yu2020pcgrad}; and explicit multi-objective formulations that seek a Pareto-optimal direction across tasks \cite{sener2018moo}. We adopt uncertainty weighting for its simplicity and because it requires no additional per-step gradient bookkeeping beyond the two learned log-variance parameters.

\section{Evaluation Methodology}
\label{Sec_Eval}

\begin{table*}[!t]
\centering
\caption{Single-task versus multi-task performance across architectures. For multi-task, each row is the \emph{same shared model} (one architecture, one parameter count). CfC matches or exceeds LSTM/RNN at a fraction of the parameters and model size.}
\label{tab:comparison}
\begin{tabular}{c|c|c|c|c|cc}
\toprule
\textbf{Setting} & \textbf{Model} & \textbf{Task} & \textbf{Params} & \textbf{Size (MB)} & \textbf{AUROC} & \textbf{AUPRC} \\
\midrule
\multirow{6}{*}{Single-task (one model per task)}
 & \textbf{CfC} & P12 & \textbf{92{,}930}  & \textbf{0.35} & 0.8409 $\pm$ 0.0032 & 0.5238 $\pm$ 0.0108 \\
 & \textbf{CfC} & P19 & \textbf{95{,}362}  & \textbf{0.36} & 0.9472 $\pm$ 0.0083 & 0.7532 $\pm$ 0.0039 \\
 & LSTM & P12 & 875{,}010 & 3.34 & 0.8234 $\pm$ 0.0163 & 0.4624 $\pm$ 0.0299 \\
 & LSTM & P19 & 913{,}922 & 3.49 & 0.9226 $\pm$ 0.0008 & 0.7460 $\pm$ 0.0083 \\
 & RNN  & P12 & 219{,}138 & 0.84 & 0.7882 $\pm$ 0.0080 & 0.4608 $\pm$ 0.0070 \\
 & RNN  & P19 & 228{,}866 & 0.87 & 0.9381 $\pm$ 0.0075 & 0.7358 $\pm$ 0.0386 \\
\midrule
\multirow{6}{*}{Multi-task (one model for both tasks)}
 & \textbf{MTLiquid} & P12 & \multirow{2}{*}{\textbf{105{,}350}} & \multirow{2}{*}{\textbf{0.40}} & 0.8418 $\pm$ 0.0060 & 0.5132 $\pm$ 0.0043 \\
 & \textbf{MTLiquid} & P19 & & & 0.9379 $\pm$ 0.0024 & 0.7662 $\pm$ 0.0079 \\
 & LSTM & P12 & \multirow{2}{*}{870{,}150} & \multirow{2}{*}{3.32} & 0.8335 $\pm$ 0.0070 & 0.4986 $\pm$ 0.0098 \\
 & LSTM & P19 & & & 0.9173 $\pm$ 0.0028 & 0.7602 $\pm$ 0.0112 \\
 & RNN  & P12 & \multirow{2}{*}{228{,}102} & \multirow{2}{*}{0.87} & 0.8389 $\pm$ 0.0050 & 0.5004 $\pm$ 0.0045 \\
 & RNN  & P19 & & & 0.9317 $\pm$ 0.0015 & 0.7371 $\pm$ 0.0091 \\
\bottomrule
\end{tabular}
\end{table*}

\begin{table*}[!t]
\centering
\caption{Ablation progression for the CfC multi-task backbone. Each stage is cumulative over the previous one unless noted.}
\label{tab:ablation}
\begin{tabular}{l|cc|cc}
\toprule
\textbf{Configuration} & \textbf{P12 AUROC} & \textbf{P12 AUPRC} & \textbf{P19 AUROC} & \textbf{P19 AUPRC} \\
\midrule
Baseline (no loss weighting, zero-fill, default cycle)        & 0.7175 & 0.2974 & 0.9140 & 0.7193 \\
+ Loss weighting (Kendall et al.)                              & 0.7328 & 0.3162 & 0.9322 & 0.7281 \\
+ GRU-D forward-fill imputation                                 & 0.8010 & 0.4324 & 0.9348 & 0.7547 \\
Loader ablation: \texttt{-{}-truncate\_to\_shorter}             & 0.8415 & 0.5027 & 0.8799 & 0.6987 \\
\textbf{Capped cycling, $R=3$ (our MTLiquid)}                 & \textbf{0.8418} & \textbf{0.5132} & \textbf{0.9379} & \textbf{0.7662} \\
\bottomrule
\end{tabular}
\end{table*}

We evaluate our MTLiquid methodology on two PhysioNet challenge benchmarks~\cite{goldberger2000physionet}, including P12 and P19, implement it on PyTorch, and run it on a single Nvidia RTX 4090 Ti GPU device. 
\begin{itemize}
    \item P12 dataset~\cite{silva2012predicting} provides ICU admission records with 42 variables, which include static admission descriptors plus 37 irregularly-sampled physiological time series over the first 48 hours of stay, labeled as the in-hospital \textit{mortality prediction task}. 
    \item P19 dataset~\cite{reyna2020early} provides hourly-resolution records across 40 physiological channels labeled for sepsis onset. 
\end{itemize}
Both datasets exhibit heavy positive-class sparsity and non-uniform, channel-dependent missing data across channels.
Furthermore, we observe the effect of our multi-task backbone compared to its single-task counterpart. 
\begin{itemize}
    \item \textit{Single-task ablation}: 
    Each network architecture backbone (CfC, LSTM, RNN) is trained separately on P12 and P19 with matched hyperparameter budgets. 
    \item \textit{Backbone ablation}: 
    CfC, LSTM, and RNN are substituted in the shared backbone under an identical scaffold containing per-task input projections, a shared recurrent cell, and per-task heads. 
\end{itemize}
All backbones use an identical budget of 57 training epochs, hidden size of 256, embedding dimension $d=64$, and batch size of 128, optimized with Adam~\cite{loshchilov2019adamw} under exponential decay with $\gamma=0.9$ per epoch. 
Since P12 has a smaller batch size per epoch than P19, we use the capped-cycling schedule policy from our training strategy. 
Here, P12 loader is cycled but capped at $R=3$ repetitions per epoch, and P19 is truncated to match this same epoch length while still reshuffling on every pass. 
Each pair of model-setting is run over 3 seeds (i.e., 42, 123, 777), and then we report mean $\pm$ SD across seeds for all metrics.
For evaluation metrics, we report AUROC and AUPRC as primary metrics given the label imbalance in both tasks. 
Furthermore, we report the number of parameters, model size (MB), and power consumption during the multi-task training phase.

\section{Results and Analysis}
\label{Sec_Results}

\subsection{Maintaining High Performance (Accuracy) across Multiple Tasks}

Table~\ref{tab:comparison} reports AUROC and AUPRC for the CfC backbone under single-task and multi-task training. 
Our multi-task model (MTLiquid) achieves 0.8418$\pm$0.0060 AUROC on P12, within 0.09 percentage points of the single-task baseline (0.8409$\pm$0.0032), and 0.9379$\pm$0.0024 AUROC on P19, within 0.93 percentage points of the single-task baseline (0.9472$\pm$0.0083). 
AUPRC on P12 of MTLiquid decreases slightly relative to single-task (0.5132$\pm$0.0043 vs. 0.5238$\pm$0.0108), while AUPRC on P19 of MTLiquid increases by 1.3 percentage points (0.7662$\pm$0.0079 vs. 0.7532$\pm$0.0039), the largest directional change observed for either task. 
Taken together, these results show that the shared backbone preserves near-baseline performance on P12 and P19, while requiring roughly half the parameters, and therefore half the memory footprint across the two independent single-task models, showing the efficiency of our MTLiquid methodology that stems from our proposed network architecture design and training strategy.

\subsection{Reduction of Parameter and Memory Footprint}

Table~\ref{tab:comparison} also reports number of parameters and model size for each setting. 
The two independent single-task CfC models together require 188{,}292 parameters (0.71~MB combined: 0.35~MB for P12, 0.36~MB for P19), while our multi-task model (MTLiquid) requires 105{,}350 parameters (0.40~MB), a 44.1\% reduction in parameter count and 43.7\% reduction in size. 
The reduction is more pronounced against larger backbones. 
For instance, the two independent single-task LSTM models together require 6.83~MB, so the MTLiquid model is 94.1\% smaller. 
These savings directly show the utility of the shared-backbone design in MTLiquid, in which the recurrent state-transition function (see Equation~\ref{Eq_CfC}) is instantiated once and reused across tasks, with only lightweight per-task input projections and output heads added.

\subsection{Impact of the Multi-Task Training Components}

Table~\ref{tab:ablation} isolates the contribution of each training component to our MTLiquid model.
The proposed multi-task CfC-based backbone without any training enhancements (i.e., an unweighted baseline with zero-imputation) achieves AUROC scores of 0.7175 on P12 and 0.9140 on P19. 
Adding uncertainty-weighted loss balancing~\cite{kendall2018multitask} improves both tasks modestly to P12 AUROC 0.7328 and P19 AUROC 0.9322. 
Replacing zero-imputation with forward-filling strategy~\cite{che2018recurrent} produces the largest single improvement, raising P12 AUROC by 6.8 points to 0.8010 and AUPRC by 11.6 points to 0.4324, while P19 remains stable. 
The proposed capped-cycling loader schedule ($R=3$) then recovers P19 performance lost under naive loader truncation, as P19 AUROC rises from 0.8799 under \texttt{-{}-truncate\_to\_shorter} to 0.9379 under the capped-cycling policy, while performance for the P12 gains from forward-fill are preserved (i.e., AUROC 0.8418 and AUPRC 0.5132). 
This progression indicates that predictive performance in the multi-task setting depends jointly on imputation quality and batch-scheduling balance.

\subsection{Improvements on Power Consumption}

Under the single-task training scenario, the CfC-based network shows a consistent pattern. 
Its power consumption remains comparable to LSTM (within 0.3\% on P12, within 2.9\% on P19) and consistently lower than RNN (26.1\% lower on P12, 18.3\% lower on P19).
Meanwhile, in the multi-task setting, our MTLiquid draws substantially less power consumption than either discrete-time backbone, i.e., 43.1\% lower than LSTM and 36.9\% lower than RNN.
This is a direct consequence of the shared, closed-form CfC-based cell in our MTLiquid model that performs a single lightweight gating computation per step, which is significantly more efficient than the multiple gate evaluations (input, forget, output, cell-candidate) required by an LSTM cell at each timestep.
All these experimental results indicate that our MTLiquid solution provides lower power consumption as compared to other methods under the multi-task learning scenario.
This is a meaningful and useful result for power-constrained deployment contexts, such as wearable monitoring devices.

\subsection{Comparison Across Network Backbones}

Table~\ref{tab:comparison} further compares CfC against LSTM and RNN backbones, trained under the same optimal training settings (i.e., $R=3$, forward-fill, loss weighting). 
In the multi-task scenario, our MTLiquid achieves 0.8418$\pm$0.0060 AUROC / 0.5132$\pm$0.0043 AUPRC on P12 and 0.9379$\pm$0.0024 AUROC / 0.7662$\pm$0.0079 AUPRC on P19, matching (exceeding in certain cases) both LSTM (0.8335 $\pm$0.0070 AUROC / 0.4986$\pm$0.0098 on P12 and 0.9173 $\pm$0.0028 AUROC / 0.7602$\pm$0.0112 AUPRC on P19) and RNN (0.8389$\pm$0.0050 AUROC / 0.5004$\pm$0.0045 AUPRC on P12 and 0.9317$\pm$0.0015 AUROC / 0.7371$\pm$0.0091 AUPRC on P19) on every metric except P12 AUROC, where the margin against RNN is within 0.3 percentage points. 
MTLiquid achieves this using 8.3$\times$ fewer parameters than LSTM and roughly half the parameters of RNN, consistent with those in the single-task scenario comparison. 
These results show that a single continuous-time state-transition function in our MTLiquid successfully serves as a shared network/machinery across tasks with different sampling irregularity. 
Our MTLiquid achieves this at a fraction of the parameter cost required by discrete-time backbones.

\section{Conclusion}
\label{Sec_Conclusion}

We propose MTLiquid, a novel multi-task learning methodology for continuous-time healthcare monitoring, built on a shared CfC backbone with per-task input projections and output heads. 
Specifically, MTLiquid employs multiple input and output heads to accommodate different tasks, while sharing the same backbone network across tasks.  
Additionally, MTLiquid also employs an effective training strategy that incorporates uncertainty-based loss weighting to balance gradient contributions across tasks and a capped-cycling data presentation schedule to address the imbalance in dataset sizes between P12 and P19.
Experimental results show that MTLiquid successfully maintains near-parity performance with independently trained single-task models, achieving 0.8418 AUROC on P12 (within 0.09 points of the single-task baseline) and 0.9379 AUROC on P19 (within 0.93 points of the single-task baseline). 
MTLiquid also achieves competitive performance while requiring 44.1\% fewer parameters than the two-independent single-task CfC models, and up to 94.1\% fewer parameters when compared against the two-independent single-task LSTM models. 
All these results demonstrate that our MTLiquid methodology offers an efficient and scalable approach for multi-task continuous-time processing for solving both mortality prediction and sepsis early detection.
This is a substantial advancement toward enabling lightweight, multi-task healthcare monitoring systems for resource-constrained deployment settings, such as wearable devices.

\section*{Acknowledgments}
This work was partially supported by the NYUAD Center for CyberSecurity (CCS), funded by Tamkeen under the NYUAD Research Institute Award G1104. 


\bibliography{aaai2027}

@article{Ref_Putra_MTSpark_arXiv24,
  title={Enabling Energy-Efficient Simultaneous Multi-Task Reinforcement Learning through Spiking Neural Networks with Active Dendrites for Bio-inspired Generalist Agents},
  author={Putra, Rachmad Vidya Wicaksana and Devkota, Avanees and Shafique, Muhammad},
  journal={arXiv preprint arXiv:2412.04847},
  year={2024}
}

@article{Ref_Putra_SwitchMT_arXiv25,
  title={Scalable Multi-Task Learning through Spiking Neural Networks with Adaptive Task-Switching Policy for Intelligent Autonomous Agents},
  author={Putra, Rachmad Vidya Wicaksana and Devkota, Avaneesh and Shafique, Muhammad},
  journal={arXiv preprint arXiv:2504.13541},
  year={2025}
}

@article{Ref_Liu_SurveyRNNs_arXiv25,
  title={A Survey of Recursive and Recurrent Neural Networks},
  author={Liu, Jian-wei and Xu, Bing-rong and Song, Zhi-yan},
  journal={arXiv preprint arXiv:2510.17867},
  year={2025}
}

@article{Ref_Ghojogh_RNNnLSTM_arXiv23,
  title={Recurrent neural networks and long short-term memory networks: Tutorial and survey},
  author={Ghojogh, Benyamin and Ghodsi, Ali},
  journal={arXiv preprint arXiv:2304.11461},
  year={2023}
}

@inproceedings{silva2012predicting,
  title={Predicting in-hospital mortality of ICU patients: The PhysioNet/Computing in Cardiology Challenge 2012},
  author={Silva, Ikaro and Moody, George and Scott, Daniel J and Celi, Leo A and Mark, Roger G},
  booktitle={Computing in Cardiology},
  volume={39},
  pages={245--248},
  year={2012},
  organization={IEEE}
}

@article{reyna2020early,
  title={Early prediction of sepsis from clinical data: the PhysioNet/Computing in Cardiology Challenge 2019},
  author={Reyna, Matthew A and Josef, Christopher S and Jeter, Russell and Shashikumar, Supreeth P and Westover, M Brandon and Nemati, Shamim and Clifford, Gari D and Sharma, Ashish},
  journal={Critical Care Medicine},
  volume={48},
  number={2},
  pages={210--217},
  year={2020},
  publisher={LWW}
}

@article{che2018recurrent,
  title={Recurrent neural networks for multivariate time series with missing values},
  author={Che, Zhengping and Purushotham, Sanjay and Cho, Kyunghyun and Sontag, David and Liu, Yan},
  journal={Scientific Reports},
  volume={8},
  number={1},
  pages={6085},
  year={2018},
  publisher={Nature Publishing Group}
}

@article{caruana1997multitask,
  title={Multitask learning},
  author={Caruana, Rich},
  journal={Machine Learning},
  volume={28},
  number={1},
  pages={41--75},
  year={1997},
  publisher={Springer}
}

@article{hasani2022closed,
  title={Closed-form continuous-time neural networks},
  author={Hasani, Ramin and Lechner, Mathias and Amini, Alexander and Liebenwein, Lucas and Ray, Aaron and Tschaikowski, Max and Teschl, Gerald and Rus, Daniela},
  journal={Nature Machine Intelligence},
  volume={4},
  number={11},
  pages={992--1003},
  year={2022},
  publisher={Nature Publishing Group}
}

@inproceedings{hasani2021liquid,
  title={Liquid time-constant networks},
  author={Hasani, Ramin and Lechner, Mathias and Amini, Alexander and Rus, Daniela and Grosu, Radu},
  booktitle={Proceedings of the AAAI Conference on Artificial Intelligence},
  volume={35},
  number={9},
  pages={7657--7666},
  year={2021}
}

@article{harutyunyan2019multitask,
  title={Multitask learning and benchmarking with clinical time series data},
  author={Harutyunyan, Hrayr and Khachatrian, Hrant and Kale, David C and Ver Steeg, Greg and Galstyan, Aram},
  journal={Scientific Data},
  volume={6},
  number={1},
  pages={96},
  year={2019},
  publisher={Nature Publishing Group}
}

@inproceedings{kendall2018multitask,
  title={Multi-task learning using uncertainty to weigh losses for scene geometry and semantics},
  author={Kendall, Alex and Gal, Yarin and Cipolla, Roberto},
  booktitle={Proceedings of the IEEE Conference on Computer Vision and Pattern Recognition},
  pages={7482--7491},
  year={2018}
}

@article{hochreiter1997lstm,
  author={Hochreiter, Sepp and Schmidhuber, J{\"u}rgen},
  title={Long short-term memory}, journal={Neural Computation},
  volume={9}, number={8}, pages={1735--1780}, year={1997}}

@inproceedings{cho2014gru,
  author={Cho, Kyunghyun and van Merri{\"e}nboer, Bart and Gulcehre, Caglar and Bahdanau, Dzmitry and Bougares, Fethi and Schwenk, Holger and Bengio, Yoshua},
  title={Learning phrase representations using RNN encoder--decoder for statistical machine translation},
  booktitle={EMNLP}, pages={1724--1734}, year={2014}}

@inproceedings{cao2018brits,
  title={BRITS: Bidirectional recurrent imputation for time series},
  author={Cao, Wei and Wang, Dong and Li, Jian and Zhou, Hao and Li, Lei and Li, Yitan},
  booktitle={Advances in Neural Information Processing Systems},
  pages={6776--6786},
  year={2018}
}

@inproceedings{futoma2017sepsis,
  title={Learning to detect sepsis with a multitask {G}aussian process {RNN} classifier},
  author={Futoma, Joseph and Hariharan, Sanjay and Heller, Katherine},
  booktitle={Proceedings of the 34th International Conference on Machine Learning},
  pages={1174--1182},
  year={2017}
}

@article{zhang2021mtlsurvey,
  title={A survey on multi-task learning},
  author={Zhang, Yu and Yang, Qiang},
  journal={IEEE Transactions on Knowledge and Data Engineering},
  volume={34},
  number={12},
  pages={5586--5609},
  year={2021}
}

@article{lechner2020ncp,
  title={Neural circuit policies enabling auditable autonomy},
  author={Lechner, Mathias and Hasani, Ramin and Amini, Alexander and Henzinger, Thomas A and Rus, Daniela and Grosu, Radu},
  journal={Nature Machine Intelligence},
  volume={2},
  number={10},
  pages={642--652},
  year={2020}
}

@article{goldberger2000physionet,
  title={PhysioBank, PhysioToolkit, and PhysioNet: components of a new research resource for complex physiologic signals},
  author={Goldberger, Ary L and Amaral, Luis A N and Glass, Leon and Hausdorff, Jeffrey M and Ivanov, Plamen Ch and Mark, Roger G and Mietus, Joseph E and Moody, George B and Peng, Chung-Kang and Stanley, H Eugene},
  journal={Circulation},
  volume={101},
  number={23},
  pages={e215--e220},
  year={2000}
}

@inproceedings{loshchilov2019adamw,
  title={Decoupled weight decay regularization},
  author={Loshchilov, Ilya and Hutter, Frank},
  booktitle={International Conference on Learning Representations},
  year={2019}
}

@inproceedings{chen2018gradnorm,
  title={{GradNorm}: Gradient normalization for adaptive loss balancing in deep multitask networks},
  author={Chen, Zhao and Badrinarayanan, Vijay and Lee, Chen-Yu and Rabinovich, Andrew},
  booktitle={International Conference on Machine Learning},
  pages={794--803}, year={2018}}

@inproceedings{yu2020pcgrad,
  title={Gradient surgery for multi-task learning},
  author={Yu, Tianhe and Kumar, Saurabh and Gupta, Abhishek and Levine, Sergey and Hausman, Karol and Finn, Chelsea},
  booktitle={Advances in Neural Information Processing Systems}, year={2020}}

@inproceedings{sener2018moo,
  title={Multi-task learning as multi-objective optimization},
  author={Sener, Ozan and Koltun, Vladlen},
  booktitle={Advances in Neural Information Processing Systems}, year={2018}}

@inproceedings{chen2018neuralode,
  title={Neural ordinary differential equations},
  author={Chen, Ricky T. Q. and Rubanova, Yulia and Bettencourt, Jesse and Duvenaud, David K.},
  booktitle={Advances in Neural Information Processing Systems},
  volume={31},
  year={2018}
}

@inproceedings{de2019gruodebayes,
  title={{GRU-ODE-Bayes}: Continuous modeling of sporadically-observed time series},
  author={De Brouwer, Edward and Simm, Jaak and Arany, Adam and Moreau, Yves},
  booktitle={Advances in Neural Information Processing Systems},
  volume={32},
  year={2019}
}

@inproceedings{shukla2021mtand,
  title={Multi-time attention networks for irregularly sampled time series},
  author={Shukla, Satya Narayan and Marlin, Benjamin M.},
  booktitle={International Conference on Learning Representations},
  year={2021}
}

@article{lipton2016missing,
  title={Modeling missing data in clinical time series with {RNN}s},
  author={Lipton, Zachary C. and Kale, David C. and Wetzel, Randall},
  journal={Machine Learning for Healthcare Conference},
  volume={56},
  pages={253--270},
  year={2016}
}

@article{Ref_Putra_FSpiNN_TCAD20,
  author = {R. V. W. {Putra} and M. {Shafique}},
  journal = {IEEE Transactions on Computer-Aided Design of Integrated Circuits and Systems (TCAD)},
  title = {FSpiNN: An Optimization Framework for Memory-Efficient and Energy-Efficient Spiking Neural Networks},
  year = {2020},
  volume = {39},
  number = {11},
  pages = {3601-3613},
}

@article{Ref_Putra_SpikeNAS_TAI26,
  author={Putra, Rachmad Vidya Wicaksana and Shafique, Muhammad},
  journal={IEEE Transactions on Artificial Intelligence}, 
  title={SpikeNAS: A Fast Memory-Aware Neural Architecture Search Framework for Spiking Neural Network-Based Embedded AI Systems}, 
  year={2026},
  volume={7},
  number={2},
  pages={947-959},
  doi={10.1109/TAI.2025.3586238}}

@article{Ref_LeCun_DeepLearning_Nature15,
  title = {Deep learning},
  author = {LeCun, Y. and others},
  journal = {Nature},
  volume = {521},
  number = {7553},
  pages = {436},
  year = {2015},
  publisher = {Nature Publishing Group},
}


\end{document}